\documentclass[a4paper,conference]{IEEEtran}
\usepackage{graphicx}
\usepackage{epsfig}
\usepackage{graphicx}
\usepackage{amsmath}
\usepackage{amssymb}
\usepackage{multirow}
\usepackage{lscape}
\usepackage{caption}
\usepackage{cite}
\usepackage{booktabs}
\usepackage[pagebackref=false,breaklinks=true,letterpaper=true,colorlinks,bookmarks=false]{hyperref}
\usepackage{makecell}

\usepackage{array}
\newcolumntype{L}[1]{>{\raggedright\let\newline\\\arraybackslash\hspace{0pt}}m{#1}}
\newcolumntype{C}[1]{>{\centering\let\newline\\\arraybackslash\hspace{0pt}}m{#1}}
\newcolumntype{R}[1]{>{\raggedleft\let\newline\\\arraybackslash\hspace{0pt}}m{#1}}

\begin{document}
%
\title{Deep Spatiotemporal Representation of the Face \\ for Automatic Pain Intensity Estimation}

\author{\IEEEauthorblockN{Mohammad Tavakolian and Abdenour Hadid}
\IEEEauthorblockA{Center for Machine Vision and Signal Analysis (CMVS)\\
University of Oulu, Finland\\
Email: {firstname.lastname}@oulu.fi
}}

\maketitle

\begin{abstract}
Automatic pain intensity assessment has a high value in disease diagnosis applications. Inspired by the fact that many diseases and brain disorders can interrupt normal facial expression formation, we aim to develop a computational model for automatic pain intensity assessment from spontaneous and micro facial variations. For this purpose, we propose a 3D deep architecture for dynamic facial video representation. The proposed model is built by stacking several convolutional modules where each module encompasses a 3D convolution kernel with a fixed temporal depth, several parallel 3D convolutional kernels with different temporal depths, and an average pooling layer. Deploying variable temporal depths in the proposed architecture allows the model to effectively capture a wide range of spatiotemporal variations on the faces. Extensive experiments on the UNBC-McMaster Shoulder Pain Expression Archive database show that our proposed model yields in a promising performance compared to the state-of-the-art in automatic pain intensity estimation. 
\end{abstract}


\IEEEpeerreviewmaketitle

\section{Introduction}
\label{Intro}
The recent medical evidences reveal that a plethora of diseases and brain disorders produce facial abnormalities and interrupt normal facial expression formation \cite{c2}. As the fifth vital sign of the health condition \cite{c3}, pain is recognized as a highly unpleasant sensation which is caused by diseases, injuries, and mental distress. So, pain is commonly considered as an indicator of the health condition. The facial expressions can be largely characterized as a reflective, spontaneous reaction to painful experiences \cite{c4}. Therefore, the detailed analysis of the facial expressions of a patient who is suffering from pain can provide objective information for practitioners for an effective disease diagnosis. 

Pain is usually reported by the patients themselves (self-report). However, this method cannot be used for the population who are incapable of articulating their pain experiences, such as neonates and unconscious people. The majority of works on pain assessment is based on the Facial Action Coding System (FACS) \cite{c5} in which the facial expressions are objectively scored in terms of elementary facial movements. These studies have mainly focused on detecting and measuring the pain intensity by extracting features from videos of the face \cite{c6,c7}. Lucy \emph{et al.} \cite{c7} trained extended SVM classifiers for three-level pain intensity estimation. Kaltwang \emph{et al.} \cite{c8} extracted LBP and DCT features from facial images and used them and their combinations as appearance-based features. They fed these features to Relevance Vector Regression (RVR) for pain intensity detection. Florea \emph{et al.} \cite{c9} improved the performance of pain intensity recognition by using a histogram of topographical features and an SVM classifier. Zhao \emph{et al.} \cite{c10} proposed an alternating direction method of multipliers to solve Ordinal Support Vector Regression (OSVR). Although traditional image descriptors represent the video frames based on static features, they are restrictive in acquiring rich dynamic information from the face. 

Recently, a few attempts have been made to model temporal information within video sequences by using deep neural networks. Zhou \emph{et al.} \cite{c11} proposed a Recurrent Convolutional Neural Network (RCNN) as a regressor model to estimate the pain intensity. They converted video frames into vectors and fed them to their model. However, this spatial conversion results in losing the structural information of the face. Rodriguez \emph{et al.} \cite{c12} extracted features of each frame from the fully connected layer of a CNN architecture. These features are fed to a Long-Short Term Memory network (LSTM) to exploit the temporal information. In this way, they consider a temporal relationship between the extracted features, while ignoring the temporal relationship among adjacent raw frames.

In order to achieve an efficient representation of the facial videos, it is crucial to capture both appearance and temporal information. In recent years, several deep models have been proposed for spatiotemporal representation. These models mainly use 3D filters with fixed temporal kernel depths. The most intuitive architecture is based on 3D Convolutional Neural Networks (CNNs) \cite{c13} where the depth of kernels corresponds to the number of frames used as input. Simonyan \emph{et al.} \cite{c14} proposed a two-stream network, including RGB spatial and optical-flow CNNs. Tran \emph{et al.} \cite{c15} explored 3D CNNs with a kernel size of $3\times 3\times 3$ to learn both the spatial and temporal features with 3D convolution operations. Sun \emph{et al.} \cite{c16} decomposed 3D convolutions into 2D spatial and 1D temporal convolutions. Carreira \emph{et al.} \cite{c17} proposed converting a pre-trained Inception-V1 \cite{c18} model to 3D by inflating all the kernels with an additional temporal dimension. They achieved this by repeating the weights of 2D filters $N$ times. All these structures have fixed temporal 3D convolution kernel depths throughout the whole architecture, that make them incapable of capturing short, mid, and long temporal ranges. 
\begin{figure*}[t]
\begin{center}
	\includegraphics[width=\linewidth]{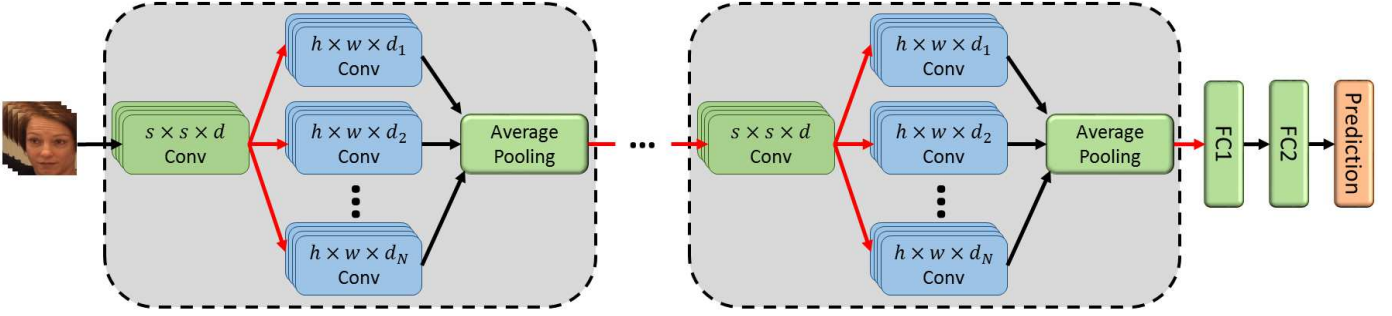}
\end{center}
	\caption{An overview of the proposed deep model. Each shaded area is a convolutional module. The red arrows show the ReLu non-linearity, while the black ones are normal connections.}
    \label{Fig1}
\end{figure*}

In this work, we strive to take a step towards the goal of automatic pain intensity estimation by developing a novel 3D convolutional model. We propose a 3D deep convolutional neural network that captures both the appearance and temporal information of videos in different temporal ranges by incorporating several temporal kernel depths. Our model learns the spatiotemporal representation of facial pain expression throughout its deep architecture and is trained end-to-end. We validate the effectiveness of our proposed model on automatic pain intensity estimation using a benchmarking and widely used database. 

\section{The Proposed Method}
\label{Method}
A subject's face, who has painful experiences, has spontaneous structural variations. In this paper, we aim to design a deep model to capture a wide range of facial dynamics in the given video sequences for an effective spatiotemporal representation. 
We develop a novel deep architecture by introducing several 3D convolutional kernels of different temporal depths to capture short, mid, and long-range variations in the whole video sequence. Figure \ref{Fig1} illustrates an overview of our proposed model. As can be seen in Figure \ref{Fig1}, the proposed model is built by stacking many convolutional modules and two fully connected layers. Each convolutional module comprises one 3D convolutional layer with the fixed temporal depth following by several parallel 3D convolutional layers with variable temporal depths. The depth of parallel 3D convolutional kernels ranges in $d\in \left\{ {{d}_{1}},{{d}_{2}},\ldots ,{{d}_{N}} \right\}$. Instead of capturing fixed temporal range homogeneously, the proposed model captures a wide range of dynamics that provide complementary information for video representation. The output of parallel 3D convolutional kernels are simply concatenated and fed to an average pooling layer.

The output feature maps of the 3D convolutions and pooling kernels at the $m$-th convolutional module extracted from an input $x\in {{R}^{H\times W\times C}}$ is a tensor $y\in {{R}^{H\times W\times {C}'}}$, where $H$ and $W$ are the spatial size of the feature maps, and $C$ and $C'$ are the number of channels of the input and output feature maps, respectively. The 3D convolutional kernels are of the size of $h\times w\times {d}_{n}$, where $h$, $w$, and ${d}_{n}$ are the height, width, and the depth of the kernel, respectively. Therefore, the output of each convolutional modules is obtained as:
\begin{equation}
y=F\left( x\left| W,\left\{ {{W}_{{{d}_{n}}}} \right\}_{n=1}^{N} \right. \right)
\end{equation}
where $W$ denotes the weight parameters of the 3D convolutional kernel with the fixed temporal kernel and $\left\{ {{W}_{{{d}_{n}}}} \right\}_{n=1}^{N}$ is a set of weight parameters for the parallel 3D convolutional kernels. It should be noted that the output of the first convolutional kernel in each module is passed through a ReLu non-linearity before feeding it to the parallel convolutional kernels (see the red arrows in Figure \ref{Fig1}). The function $F\left( \cdot  \right)$ represents all the convolution and average pooling operations that are done in each module and defined as:
\begin{equation}
F\left( x \right)=\bar{p}\left( \bigcup\limits_{n=1}^{N}{{{W}_{{{d}_{n}}}}}\sigma \left( Wx \right) \right)
\end{equation}
where $\sigma$ denotes the ReLu non-linearty and $\bigcup{{}}$ stands for concatenation operation. $\bar{p}$ is the average pooling operator.

Within each convolutional module, after convolving the feature map of the preceding layer $x$ with the first convolution kernel, $N$ intermediate feature maps $\left\{ {{S}_{1}},{{S}_{2}},\ldots ,{{S}_{N}} \right\}$ are obtained, where ${{S}_{n}}\in {{R}^{H\times W\times {{C}_{n}}}}$. Each intermediate feature map has a different number of channels as they are obtained by convolution operations of different temporal depths, while the spatial size of all feature maps $H\times W$ is the same. These intermediate feature maps $\left\{ {{S}_{n}} \right\}_{n=1}^{N}$ are simply concatenated into a single 3D tensor and then fed into an average pooling layer. As shown in Figure \ref{Fig1}, the entire of the model are trained in an end-to-end network training basis.

For pain intensity estimation, the model should make a continuous-valued prediction. Therefore, instead of cross-entropy loss function which is frequently used for classification in deep architectures, we use the mean squared error function to turn it into a regression problem. We calculate the Euclidean distance between the predicted output $\hat{o}$ and the actual one $o$ to obtain the error. The training is carried out using Stochastic Gradient Descent (SGD) and the backpropagation.
\begin{equation}
E=\frac{1}{K}\sum\limits_{k=1}^{K}{\left\| {{{\hat{o}}}_{k}}-{{o}_{k}} \right\|_{2}^{2}}
\end{equation}
where $K$ is the total number of predictions for the given video sequence.

\section{Experimental Results}
\begin{figure}[t]
\begin{center}
	\includegraphics[width=\linewidth]{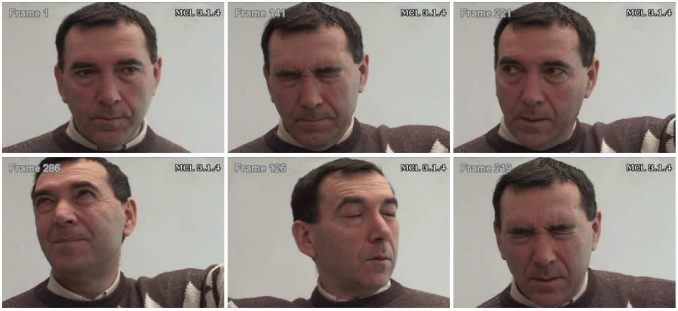}
\end{center}
	\caption{Example frames from the UNBC-McMaster Shoulder Pain Expression Archive database \cite{c1}.}
    \label{UNBC}
\end{figure}

In this section, we perform a series of extensive comparative experiments to evaluate the effectiveness of the proposed model for automatic pain intensity estimation on the widely used UNBC-McMaster Shoulder Pain Expression Archive database \cite{c1}. The database includes facial videos of individuals who were performing a series of active and passive range-of-motion tests to their either affected or unaffected limbs on two sessions. Figure \ref{UNBC} shows some samples from this database. Each video sequence was annotated in a frame-level fashion by FACS, resulting in 16 discrete pain intensity levels (0-15) based on facial Action Units (AUs). In our experiments, we used the active tests set that includes 200 videos of 25 subjects with 48,398 frames of the size $320\times 240$ pixels.

First, we performed an inclusive search to find the optimal values for the model's hyper-parameters. Then, we drew a comparison between the performance of the proposed method and the state-of-the-art approaches. In our experiments, we followed the standard protocol as \cite{c11,c12} to evaluate the performance of the proposed model. 

\subsection{Implementation Setup}
\begin{figure}[t]
\begin{center}
	\includegraphics[width=\linewidth]{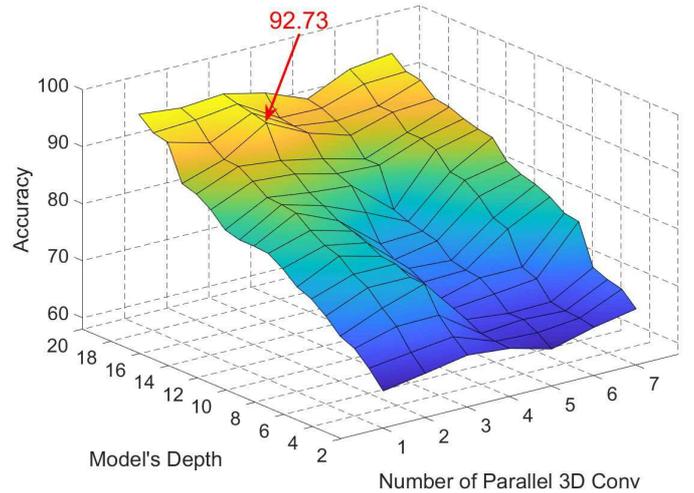}
\end{center}
	\caption{The accuracy of the proposed method versus the model's depth and the number of parallel 3D convolution kernels.}
    \label{Accuracy}
\end{figure}

We adjust the appropriate values for the hyper-parameters by performing a grid search and following the guidelines in \cite{c19}. We use the Area Under the Curve (AUC) as the accuracy metric. We build our model by stacking multiple consecutive convolutional modules (see Section \ref{Method}). In each architecture, we change the number of the parallel 3D convolution kernels to determine the appropriate temporal range that is required to be captured. Figure \ref{Accuracy} shows the normalized accuracy of the proposed method for different network sizes and the number of 3D parallel convolution kernels. It can be seen that the highest accuracy is obtained when the number of parallel 3D convolutions and the model's depth are set to 3 and 17, respectively. Therefore, our model has 53 layers including convolution, average pooling, and fully connected layers. In our framework, the parallel 3D convolutions of each module are considered as one layer, as their input is the same and their output feature maps are concatenated into a single tensor.

In addition, the size of the receptive field of the kernels determines the amount of the information that the model captures. Proper kernel's structure plays a crucial role in capturing detailed information for an effective representation. Hence, we conducted some experiments to find the optimal size of the receptive field. Table \ref{Table1} reports the accuracy of the proposed model on a range of receptive field’s sizes. The highest accuracy belongs to the model with the spatial kernel size of $3\times 3$. From these results, we can conclude that the larger the spatial size of the kernels, coarser information is captured, resulting in poor video representation in this framework.

\begin{table}[t]
\centering
\caption{Comparison of the proposed model's accuracy (\%) for different spatial sizes of the kernels’ receptive field.}
\label{Table1}
\begin{tabular}{L{1.8cm}C{0.8cm}C{0.8cm}C{0.8cm}C{0.8cm}C{1.1cm}}
\toprule
\textbf{Spatial Size} & $\mathbf{3\times 3}$ & $\mathbf{5\times 5}$ & $\mathbf{7\times 7}$ & $\mathbf{9\times 9}$ & $\mathbf{11\times 11}$ \\ \noalign{\smallskip} \hline \noalign{\smallskip}
\textbf{Accuracy}     & 92.73        & 89.31        & 84.05        & 79.43        & 72.18         \\ \bottomrule
\end{tabular}
\end{table}

According to \cite{c15}, the temporal depth of input data to model influences the model performance in spatiotemporal representation. So, we evaluate the accuracy of our model for inputs of different temporal depths (see Table \ref{Table2}). Among all the tested temporal depths, the input data with the temporal depth of 32 frames performs better. This result partially validates our initial hypotheses that larger input depth allows the model to capture short, mid, and long-range spatiotemporal terms in the video for more efficient representation.

\begin{table}[t]
\centering
\caption{Comparison of the proposed model's accuracy (\%) on the different temporal depth of the input data.}
\label{Table2}
\begin{tabular}{L{2cm}C{0.65cm}C{0.65cm}C{0.65cm}C{0.65cm}C{0.65cm}C{0.65cm}}
\toprule
\textbf{Temporal Depth} & \textbf{3} & \textbf{5} & \textbf{8} & \textbf{16} & \textbf{32} & \textbf{40} \\ \noalign{\smallskip} \hline \noalign{\smallskip}
\textbf{Accuracy}       & 61.04       & 79.55       & 84.39       & 90.41        & 92.73        & 87.36        \\ \bottomrule
\end{tabular}
\end{table}

Based on the achieved results in this section, the proposed model reaches its optimal performance with 17 convolutional modules where each of them has 3 parallel 3D kernels with the spatial receptive field of the size $3\times 3$. Moreover, the temporal depth of the input is set to 32 frames. We use these settings in all the remaining experiments. For training the model, we conform to the weight initialization strategy in \cite{c20}. Moreover, we use Stochastic Gradient Descent (SGD) with a momentum of 0.9, weight decay of $10^{-4}$, and batch size of 64. The initial learning rate is set to 0.01 and reduced by a factor of 10 after every 10 epochs. The maximum number of epochs is 150.  

\subsection{Comparison with the state-of-the-art}
In this section, we evaluate the effectiveness of the proposed model by comparing its performance with the recent state-of-the-art methods in automatic pain intensity estimation on the UNBC-McMaster database \cite{c1}. In order to make a fair comparison with the benchmark methods, we report the Mean Squared Error (MSE), the Pearson Correlation Coefficient (PCC), and the Intra-Class Correlation Coefficient (ICC). 

Table \ref{Table3} compares the proposed method with benchmark approaches.  The proposed model shows a promising performance compared to the state-of-the-art methods. As can be seen, the proposed method consistently outperforms the existing benchmark approaches by a large margin both in terms of MSE, PCC, and ICC. It significantly reduces the MSE by 0.18 compared to the basic 3D CNN \cite{c15}. In addition, its high PCC reveals that the proposed method is able to effectively extract detailed information from the face structure for pain intensity estimation. Moreover, the obtained high ICC demonstrates strong resemblance in each level of pain despite the subject's identity in the leave-one-subject-out cross-validation. 
\begin{table}[t]
\centering
\caption{Comparison of the Mean Squared Error, the Pearson Correlation Coefficient, and the Intra-Class Correlation Coefficient on the UNBC-McMaster database \cite{c1}.}
\label{Table3}
\begin{tabular}{L{3.6cm}C{1.15cm}C{1.15cm}C{1.15cm}}
\toprule
                            & \textbf{MSE}  & \textbf{PCC}  & \textbf{ICC}  \\ \noalign{\smallskip} \hline \hline \noalign{\smallskip}
Kaltwang \emph{et al.} \cite{c8}     & 1.39          & 0.59          & 0.50          \\ \noalign{\smallskip} \hline \noalign{\smallskip}
Florea \emph{et al.} \cite{c9}        & 1.21          & 0.53          & N/A           \\ \noalign{\smallskip} \hline \noalign{\smallskip}
Zhao \emph{et al.} \cite{c10}        & N/A           & 0.60          & 0.56          \\ \noalign{\smallskip} \hline \noalign{\smallskip}
Zhou \emph{et al.} \cite{c11}        & 1.54          & 0.64          & N/A           \\ \noalign{\smallskip} \hline \noalign{\smallskip}
Rodriguez \emph{et al.} \cite{c12}   & 0.74          & 0.78          & 0.45          \\ \noalign{\smallskip} \hline \noalign{\smallskip}
Tran \emph{et al.} \cite{c15}\textsuperscript{\dag}        & 0.71          & 0.81          & 0.41          \\ \noalign{\smallskip} \hline \noalign{\smallskip}
\textbf{The Proposed Model} & \textbf{0.53} & \textbf{0.84} & \textbf{0.75} \\ \bottomrule 
\multicolumn{4}{p{8.4cm}}{\textsuperscript{\dag}{\makecell{\footnotesize The method was originally proposed for action recognition and is adopt-\\ed for pain intensity estimation.}}
}
\end{tabular}
\end{table}

\begin{figure}[t]
\begin{center}
	\includegraphics[width=\linewidth]{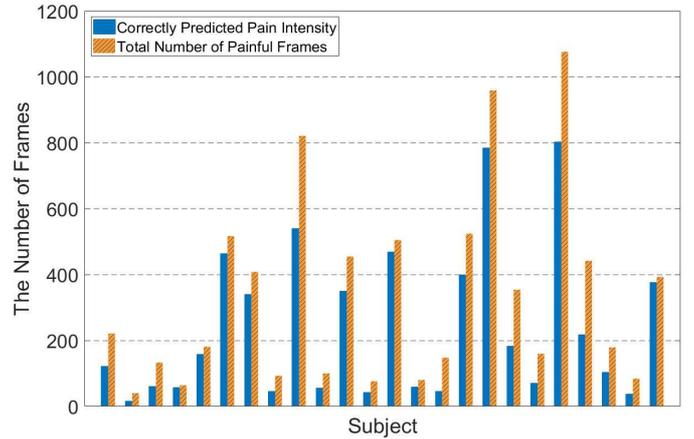}
\end{center}
	\caption{The correctly pain intensity level versus the total number of painful frames for each subject.}
    \label{Prediction}
\end{figure}

To gain better insights into the effectiveness of the proposed method, Figure \ref{Prediction} shows the number of frames whose pain intensity level has been correctly predicted versus the total number of frames for each subject.

\section{Conclusion}
Automatic pain assessment plays a crucial role in the healthcare applications, especially disease diagnosis. This paper introduced a novel deep architecture to represent the facial video for pain intensity estimation. The proposed model extracts the detailed spatiotemporal information of spontaneous variations in the face expression by exploiting several parallel 3D convolution operations with divers temporal depths. Unlike 3D convolutional neural networks with fixed 3D homogeneous kernel depths, our proposed architecture captures short, mid, and long-range spatiotemporal variations that allow representing the spontaneous facial variations more effectively. Feeding the obtained representation of the given video to a regressor, we estimate the level of pain intensity. Our extensive experiments on the UNBC-McMaster Shoulder Pain Expression Archive database showed the effectiveness of the proposed method in the automatic pain intensity estimation. As a future work, we plan to explore the generalization of our model on other face analysis tasks.

\bibliographystyle{IEEEtran}
\bibliography{Ref_ICPR18}

\begin{thebibliography}{10}
\providecommand{\url}[1]{#1}
\csname url@samestyle\endcsname
\providecommand{\newblock}{\relax}
\providecommand{\bibinfo}[2]{#2}
\providecommand{\BIBentrySTDinterwordspacing}{\spaceskip=0pt\relax}
\providecommand{\BIBentryALTinterwordstretchfactor}{4}
\providecommand{\BIBentryALTinterwordspacing}{\spaceskip=\fontdimen2\font plus
\BIBentryALTinterwordstretchfactor\fontdimen3\font minus
  \fontdimen4\font\relax}
\providecommand{\BIBforeignlanguage}[2]{{%
\expandafter\ifx\csname l@#1\endcsname\relax
\typeout{** WARNING: IEEEtran.bst: No hyphenation pattern has been}%
\typeout{** loaded for the language `#1'. Using the pattern for}%
\typeout{** the default language instead.}%
\else
\language=\csname l@#1\endcsname
\fi
#2}}
\providecommand{\BIBdecl}{\relax}
\BIBdecl

\bibitem{c2}
P.~H. Tseng, I.~G.~M. Cameron, G.~Pari, J.~N. Reynolds, D.~P. Munoz, and
  L.~Itti, ``High-throughput classification of clinical populations from
  natural viewing eye movements,'' \emph{Journal of Neurology}, vol. 260,
  no.~1, pp. 275--284, 2013.

\bibitem{c3}
M.~Lynch, ``Pain as the fifth vital sign,'' \emph{Journal of Intravenous
  Nursing}, vol.~24, no.~2, 2001.

\bibitem{c4}
K.~D. Craig, K.~M. Prkachin, and R.~V.~E. Grunau, \emph{Handbook of Pain
  Assessment}.\hskip 1em plus 0.5em minus 0.4em\relax Guilford Press, 2011, ch.
  The Facial Expression of Pain.

\bibitem{c5}
P.~Ekman and W.~Friesen, \emph{Facial Action Coding System: A Technique for the
  Measurement of Facial Movement}.\hskip 1em plus 0.5em minus 0.4em\relax
  Consulting Psychologists Press, 1978.

\bibitem{c6}
S.~Brahnam, C.~F. Chuang, F.~Y. Shih, and M.~R. Slack, ``Machine recognition
  and representation of neonatal facial displays of acute pain,''
  \emph{Artificial Intelligence in Medicine}, vol.~36, no.~3, pp. 211--222,
  2006.

\bibitem{c7}
P.~Lucey, J.~F. Cohn, K.~M. Prkachin, P.~E. Solomon, S.~Chew, and I.~Matthews,
  ``Painful monitoring: Automatic pain monitoring using the unbc-mcmaster
  shoulder pain expression archive database,'' \emph{Image and Vision
  Computing}, vol.~30, no.~3, pp. 197--205, 2012.

\bibitem{c8}
S.~Kaltwang, O.~Rudovic, and M.~Pantic, ``Continuous pain intensity estimation
  from facial expressions,'' in \emph{Int. Symp. Advances in Visual Computing},
  2012, pp. 368--377.

\bibitem{c9}
C.~Florea, L.~Florea, and C.~Vertan, ``Learning pain from emotion: Transferred
  {HoT} data representation for pain intensity estimation,'' in \emph{ECCV
  Workshops}, 2014, pp. 778--790.

\bibitem{c10}
R.~Zhao, Q.~Gan, S.~Wang, and Q.~Ji, ``Facial expression intensity estimation
  using ordinal information,'' in \emph{IEEE CVPR}, 2016, pp. 3466--3474.

\bibitem{c11}
J.~Zhou, X.~Hong, F.~Su, and G.~Zhao, ``Recurrent convolutional neural network
  regression for continuous pain intensity estimation in video,'' in \emph{IEEE
  CVPR Workshops}, 2016, pp. 1535--1543.

\bibitem{c12}
P.~Rodriguez, G.~Cucurull, J.~Gonzàlez, J.~M. Gonfaus, K.~Nasrollahi, T.~B.
  Moeslund, and F.~X. Roca, ``Deep pain: Exploiting long short-term memory
  networks for facial expression classification,'' \emph{IEEE Trans.
  Cybernetics}, vol.~PP, no.~99, pp. 1--11, 2017.

\bibitem{c13}
S.~Ji, W.~Xu, M.~Yang, and K.~Yu, ``{3D} convolutional neural networks for
  human action recognition,'' \emph{IEEE Trans. PAMI}, vol.~35, no.~1, pp.
  221--231, 2013.

\bibitem{c14}
K.~Simonyan and A.~Zisserman, ``Two-stream convolutional networks for action
  recognition in videos,'' in \emph{NIPS}, 2014, pp. 568--576.

\bibitem{c15}
D.~Tran, L.~Bourdev, R.~Fergus, L.~Torresani, and M.~Paluri, ``Learning
  spatiotemporal features with 3d convolutional networks,'' in \emph{IEEE
  ICCV}, 2015, pp. 4489--4497.

\bibitem{c16}
L.~Sun, K.~Jia, D.~Y. Yeung, and B.~E. Shi, ``Human action recognition using
  factorized spatio-temporal convolutional networks,'' in \emph{IEEE ICCV},
  2015, pp. 4597--4605.

\bibitem{c17}
J.~Carreira and A.~Zisserman, ``Quo vadis, action recognition? a new model and
  the kinetics dataset,'' in \emph{IEEE CVPR}, 2017, pp. 4724--4733.

\bibitem{c18}
S.~Ioffe and C.~Szegedy, ``Batch normalization: Accelerating deep network
  training by reducing internal covariate shift,'' in \emph{ICML}, 2015, pp.
  448--456.

\bibitem{c1}
P.~Lucey, J.~F. Cohn, K.~M. Prkachin, P.~E. Solomon, and I.~Matthews, ``Painful
  data: The unbc-mcmaster shoulder pain expression archive database,'' in
  \emph{IEEE Int. Conf. Face and Gesture}, 2011, pp. 57--64.

\bibitem{c19}
Y.~Bengio, \emph{Practical Recommendations for Gradient-Based Training of Deep
  Architectures}.\hskip 1em plus 0.5em minus 0.4em\relax Springer Berlin
  Heidelberg, 2012, pp. 437--478.

\bibitem{c20}
K.~He, X.~Zhang, S.~Ren, and J.~Sun, ``Delving deep into rectifiers: Surpassing
  human-level performance on imagenet classification,'' in \emph{IEEE ICCV},
  2015, pp. 1026--1034.

\end{thebibliography}




\end{document}